\documentclass[10pt,twocolumn,letterpaper]{article}

\usepackage[pagenumbers]{cvpr} % To force page numbers, e.g. for an arXiv version

\usepackage{graphicx}
\usepackage{amsmath}
\usepackage{amssymb}
\usepackage{booktabs}
\usepackage{fontawesome}

\usepackage[accsupp]{axessibility}

\usepackage{float}

\usepackage[pagebackref,breaklinks,colorlinks]{hyperref}

\usepackage[capitalize]{cleveref}
\crefname{section}{Sec.}{Secs.}
\Crefname{section}{Section}{Sections}
\Crefname{table}{Table}{Tables}
\crefname{table}{Tab.}{Tabs.}

\def\cvprPaperID{xxx} % *** Enter the CVPR Paper ID here
\def\confName{CVPR}
\def\confYear{2024}

\newcommand{\noise}{\epsilon}
\newcommand{\pnoise}{\hat{\noise}}

\newcommand{\rpose}{p'}

\begin{document}

\title{ID-Pose: Sparse-view Camera Pose Estimation by Inverting Diffusion Models}

\author{Weihao Cheng \quad Yan-Pei Cao \quad Ying Shan\\
ARC Lab, Tencent PCG \\
{\tt\small whcheng@tencent.com \quad caoyanpei@gmail.com \quad yingsshan@tencent.com}
% \and
}

\twocolumn[{
\maketitle
\vspace{-0.7cm}
\begin{figure}[H]
    \hsize=\textwidth
    \centering
    \includegraphics[width=\textwidth]{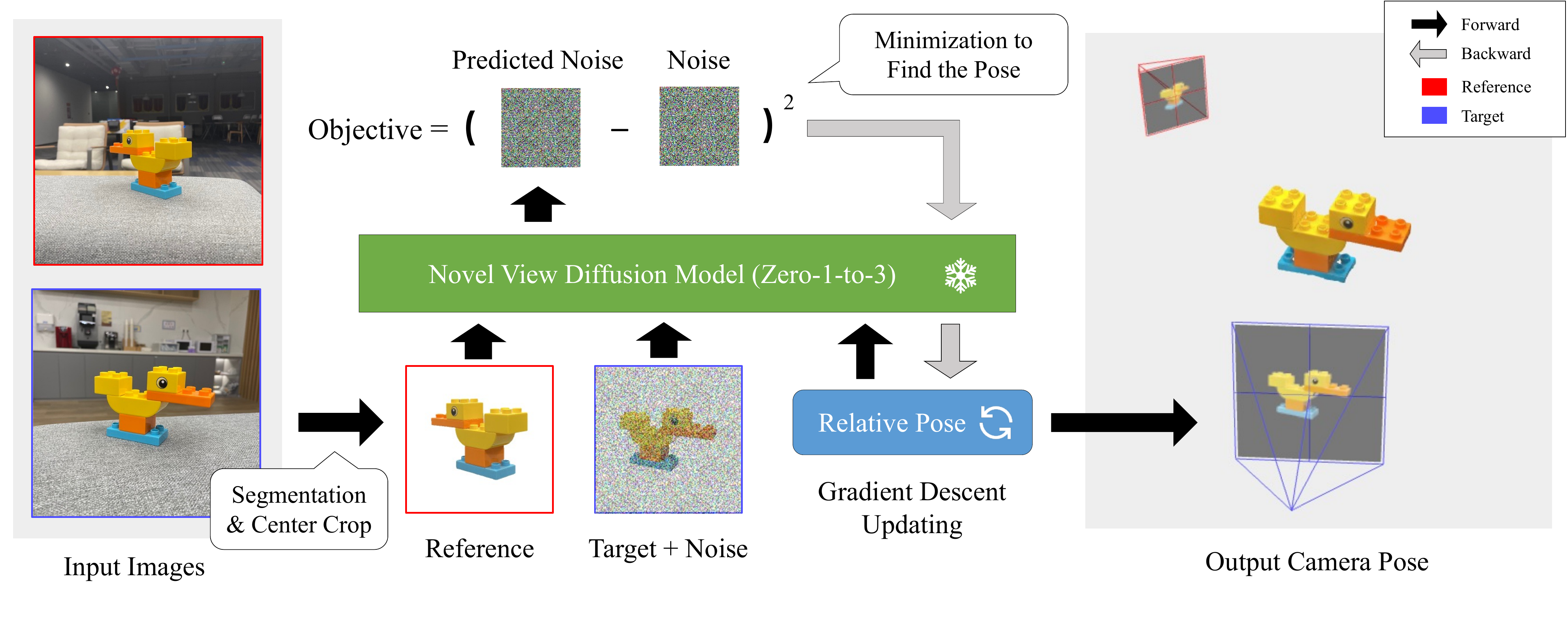}
    \caption{ID-Pose estimates the relative camera pose of two input images. It uses one image as the reference and the other image as the target, where the target image is added with Gaussian noise. Assume there is a relative transformation from the camera pose of the reference to the target. ID-Pose utilizes the noise predictor of Zero-1-to-3 \cite{zero123} to compute the noise given the noisy target image, the reference image, and the relative pose. The prediction error is used as the minimization objective regarding the relative pose. ID-Pose then applies the gradient descent method to find the optimal relative pose that connects the two views.}
    \label{fig:teaser}
\end{figure}
\vspace{1.2cm}
}]

%%%%%%%%% ABSTRACT
\begin{abstract}
Given sparse views of a 3D object, estimating their camera poses is a long-standing and intractable problem. Toward this goal, we consider harnessing the pre-trained diffusion model of novel views conditioned on viewpoints (Zero-1-to-3). We present ID-Pose which inverses the denoising diffusion process to estimate the relative pose given two input images. ID-Pose adds a noise to one image, and predicts the noise conditioned on the other image and a hypothesis of the relative pose. The prediction error is used as the minimization objective to find the optimal pose with the gradient descent method. We extend ID-Pose to handle more than two images and estimate each pose with multiple image pairs from triangular relations. ID-Pose requires no training and generalizes to open-world images. We conduct extensive experiments using casually captured photos and rendered images with random viewpoints. The results demonstrate that ID-Pose significantly outperforms state-of-the-art methods. [Project Page: \url{https://xt4d.github.io/id-pose-web/}]

\end{abstract}

%%%%%%%%% BODY TEXT
\section{Introduction}
\label{sec:intro}
Estimating camera poses of images that depict a 3D object is important to shape understanding \cite{mvdepthnet,dbw}, reconstruction \cite{nerf,pixelnerf,forge,sparsefusion}, and generation \cite{srt, viewformer, c123, one2345}. With densely scanned images, this problem is well-addressed with Structural-from-Motion (SfM) \cite{sfm}. However, given casually captured images, which are often sparse in terms of viewpoints, SfM methods frequently encounter convergence difficulties due to the insufficient overlapping appearances between different views. With a growing interest in unveiling 3D from daily images, addressing the challenge of estimating camera poses in sparse-view settings becomes an increasingly urgent task.

Recent methods train feed-forward neural networks to predict camera poses given images \cite{relative,relpose,sparsepose,relpose++}. These methods require training with accurate pose annotations, which are usually obtained by running SfM on densely captured images \cite{co3d} or human labeling \cite{navi}. As it is expensive to collect such dense data at large scale, the generalization ability of these methods is often limited. In fact, current experiments are primarily conducted on turntable-like videos of common objects \cite{co3d}, which exhibit a relatively simplistic appearance and viewpoint distributions in comparison to that of open-world images.

We solve the sparse-view camera pose estimation problem by harnessing the pre-trained diffusion model of novel views conditioned on viewpoints (Zero-1-to-3) \cite{zero123}.
We present ID-Pose which inverses the denoising diffusion process to estimate the relative pose of two input images. ID-Pose leverages the noise predictor of the diffusion model, and maintains the relative pose as the decision variable. The predictor conditions on one image and the current value of the pose to predict the noise that is added to the latent of the other image. The error of the prediction is used as the objective to update the pose with the gradient descent method. 
To improve the stability, we propose a two-stage strategy including an exploration process that finds a promising initial pose, and then a refinement process that further enhances the accuracy of the pose. 
The exploration process first samples a number of candidature poses with a few rounds of optimization and then selects the one with the minimum prediction error of noise. 
The refinement process keeps updating the selected pose until the stopping criteria are satisfied.
We further extend ID-Pose to effectively find relative poses of more than two images. The idea is to estimate a pose with multiple image pairs. Within a group of three images, we can use the poses of two image pairs to present the pose of the third pair. Therefore, the prediction error of the third pair can be used to update the two poses. This triangular relation enables the incorporation of additional information, thereby improving the accuracy of the estimation.
In general, ID-Pose transfers the image generation ability from the large-scale model to the camera pose estimation ability.
The advantages of the method are as follows:
\begin{itemize}
    \item It is a zero-shot method that requires no training;
    \item Generalizes to open-world images with varying object categories, image styles, and viewpoints;
    \item Based on the understanding of objects, so that appearance overlaps are not required in input views.
\end{itemize}
We conduct extensive experiments on NAVI \cite{navi}, OmniObject3D \cite{omniobject3d}, ABO \cite{abo}, and CO3D \cite{co3d} datasets, which include both casually captured photos and rendered images with randomly sampled viewpoints. The results demonstrate that ID-Pose outperforms state-of-the-art methods with strong generalization capability.

\section{Related Work}
\label{sec:related}

\textbf{Camera Pose Estimation.} Estimating the camera poses of images is a fundamental problem in 3D computer vision. Traditional methods \cite{sfm} extract local features \cite{sift,daisy,surf} of input images and find feature correspondences to estimate the camera poses. When images are sparsely captured, they often fail due to insufficient feature matches. Recent learning-based methods \cite{relative,relpose,sparsepose,relpose++} use neural networks to predict the relative poses of image pairs. These methods allow sparse-view input, but the training requires a large number of pose annotations, which are usually obtained from dense views \cite{co3d}. As training data is expensive to acquire, the neural networks are limited to benchmarks and hard to handle out-of-distribution data.

\textbf{Image Diffusion Models.} Diffusion models \cite{diffusion} have been widely used for image generation tasks. These models learn a neural network to denoise a noisy image, that predicts the noise value added to the image and subsequently subtracts it. By progressive denoising from pure noise, a clear and complex image can be generated. Latent diffusion models \cite{latentdiffusion} denoise latent representations of images, which improves efficiency and stability. Recent models are able to control the generation conditioned on other inputs, for example, natural language and sketches \cite{controlnet,t2i}. Stable Diffusion \cite{latentdiffusion} is a latent diffusion model that generates high-quality images based on input texts by training on over a hundred million image-text pairs. A view-conditioned diffusion model, Zero-1-to-3 \cite{zero123}, uses an input image of an object and a relative pose to generate a novel view of the object under the new viewpoint. It is fine-tuned from Stable Diffusion with pose-annotated images rendered from the Objaverse dataset \cite{objaverse}, and achieves prominent performance on real-world images. Given that Zero-1-to-3 associates views with viewpoints, we explore the possibility of utilizing this model to reveal the relative camera pose between two images. This will enable zero-shot camera pose estimation in the context of sparse views with small or no overlapping appearance.

\textbf{Inversion Techniques} Given a feed-forward model that maps an input to the corresponding output, inversion techniques convert the output back to the input using optimization. In image editing, inversion is commonly used to convert an image to a latent vector, and one can edit the vector to modify the image \cite{junyan,ganinversion}. By using Stable Diffusion, TextualInversion \cite{textualinversion} inverts images to text embedding that enables the generation of new images with the same concepts or objects. iNeRF \cite{inerf} showcases that the camera pose of an image can be inversely found conditioned on a neural 3D scene \cite{nerf}. Inspired by these approaches, we investigate inversely using novel-view generative models to estimate relative camera poses of given images, through optimization-based techniques.

\section{Methodology}
Our task is to estimate camera poses of images that are sparsely captured around an object. We present ID-Pose, a zero-shot method that takes two images as input and finds their relative camera pose. The idea is to invert the denoising process of the pre-trained novel-view diffusion model, i.e., Zero-1-to-3 \cite{zero123}, where the error of the noise prediction is minimized to obtain the pose. We further extend ID-Pose to handle more than two images by inverting each of the poses with multiple image pairs. In this section, we first introduce the background knowledge of diffusion models, and then describe the details of the ID-Pose method.

\subsection{Preliminary}
Let $x$ be an image, and $z$ be the latent map of $x$. Given a Gaussian noise $\noise$, one can add a small portion of $\noise$ to $z$ step-by-step, and finally turn $z$ into $\noise$. At each step $t$, a noisy latent map $z(\noise, t)$ is obtained by mixing $z$ and $\noise$ with proportion scales controlled by $t$. Latent diffusion models \cite{latentdiffusion} learn a step-wise noise predictor $f(z(\noise, t), t)$ that predicts the noise added to $z(\noise, t)$. To generate an image, a latent diffusion model starts from a noise $\noise$ and assumes there is a $z$ underlying the $\noise$. It then predicts and removes the noise step-by-step. After a fixed number of steps, it recovers the latent map $z$, which is then input to an image decoder to obtain the resulting image.

Stable Diffusion \cite{latentdiffusion} is a latent diffusion model that generates images conditioned on text. Its noise predictor is therefore written as $f(z(\noise, t), text, t)$. The condition $text$ is mapped to CLIP \cite{clip} embedding tokens which are cross-attended with the hidden states of the predictor network. 

Zero-1-to-3 \cite{zero123} is fine-tuned from Stable Diffusion that generates novel views of an object conditioned on a reference view $x_c$ and a relative camera pose $p$. The noise predictor of Zero-1-to-3 is written as $f(z(\noise, t), x_c, p, t)$. For convenient representations, we simplify this notation to $f(z(\noise), x, p)$ where $t$ is omitted. The reference view $x_c$ is mapped to image tokens with the CLIP image encoder. The camera pose $p$ is expressed in the spherical coordinate system where the position of a point is specified by three numbers: polar angle, azimuth angle, and radius. The three values of the pose $p$ are concatenated with the $x_c$ tokens as conditions to the predictor $f$. Zero-1-to-3 is fine-tuned with large-scale images rendered from 3D objects \cite{objaverse} but generalizes to real-world images. 

\subsection{ID-Pose}
Recall that $f(z(\noise), x, p)$ is the pre-trained noise predictor of Zero-1-to-3.
Given two images $x_0$ and $x_1$, we want to use $f$ to find their relative pose $p$ that is represented in spherical coordinates (polar, azimuth, radius). 
Assume $x_0$ is the reference view and $x_1$ is the target view. We can add a noise $\noise$ to $x_1$ and use $f$ to predict that noise. We define $p$ as the pose transformation from $x_0$ to $x_1$, the predicted noise can be considered as a function of $p$:
\begin{equation}
    \pnoise(p \,;\, x_0, x_1) = f(z_1(\noise), x_0, p),
\end{equation}
where $z_1(\noise)$ is the noisy latent map of $x_1$. We define the noise error of the prediction given $p$ as:
\begin{equation}\label{eq:error}
    l(p \,;\, x_0, x_1) = || \pnoise(p \,;\, x_0, x_1) - \noise ||_2^2.
\end{equation}
We also consider that $x_1$ acts as the reference view and $x_0$ acts as the target view. Let $\rpose$ be the reversed transformation of $p$ (from $x_1$ to $x_0$), where $\rpose = -p$ is held in spherical coordinates. The predicted noise regarding $\rpose$ is defined as:
\begin{equation}
    \pnoise(\rpose \,;\,x_1, x_0) = f(z_0(\noise), x_1, \rpose),
\end{equation}
where $z_0(\noise)$ is the noisy latent map of $x_0$. Thereby, the noise error of the prediction given $\rpose$ is:
\begin{equation}
    l(\rpose \,;\, x_1, x_0) = || \pnoise(\rpose \,;\, x_1, x_0) - \noise ||_2^2.
\end{equation}
We can inversely find $p$ by minimizing either of the noise errors:
\begin{equation}
    \min_{p}{ l(p \,;\, x_0, x_1) \; \text{or} \; l(\rpose \,;\, x_1, x_0) }.
\end{equation}
We alternately choose $l(p \,;\, x_0, x_1)$ or $l(\rpose \,;\, x_1, x_0)$ as the objective function $g(p)$. The optimization process iteratively updates $p$ using the gradient descent method:
\begin{equation}\label{eq:gd}
    p^{(k+1)} = p^{(k)} - \alpha \nabla g(p^{(k)}), %\frac{\partial g(p^{(k)}) }{\partial p},
\end{equation}
where $\alpha$ is the updating factor, $k$ represents the current iteration, and $k+1$ represents the next iteration. The optimization stops when $p$ converges or $k$ exceeds a preset maximum count. The intuition is that, $p$ reaches the optimum once the model $f$ predicts the noise with the minimum error.

The initial value of $p$ is crucial as optimization can be often stuck into local minimums. In the case of nearly symmetric objects, the opposing view of a targeting view may exhibit the greatest similarity when compared to the other views. Under this situation, if $p$ is initialized near this opposing view, it will be likely stuck into a local minimum. 
To alleviate this problem, we design ID-Pose with two stages. The first stage runs an exploration process to find a reliable initial value of $p$. The second stage runs a refinement process to further update $p$ with Equation \eqref{eq:gd} until stops. We describe the exploration process as follows. We first create $m$ candidature relative poses, where the azimuth angles are uniformly sampled from $0$ to $2\pi$. We then use the method from \cite{one2345} to estimate the absolute elevations of the two images, where the difference is used to set the relative polar angles of the candidates. We lastly set the relative radius to zero. 
For each candidature pose, we update it for a few iterations with Equation \eqref{eq:gd}. We then probe those poses with the \textit{pairwise error}:
\begin{equation}\label{eq:probe}
    l_{0,1}(p) = l(p \,;\, x_0, x_1) + l(\rpose \,;\, x_1, x_0),
\end{equation}
which combines the two noise errors to measure the instability of $p$ connecting $x_0$ and $x_1$. From all the candidates, we pick the one with the minimum \textit{pairwise error} as the initial pose.

\subsection{ID-Pose with More Views}
Given more than two images, i.e., $n > 2$, a naive strategy is selecting an anchor image and creating $n-1$ image pairs, where each pose is inverted independently. However, there are many connectable but unused relations that can help find these poses jointly. We consequently consider estimating each of the relative poses with multiple image pairs from triangular relations. Suppose the input images are $x_0$, $x_1$, ..., $x_{n-1}$, we maintain $n-1$ relative poses $p_1$, $p_2$, ..., $p_{n-1}$ as variables, where $p_i$ represents the relative transformation from $x_0$ to $x_i$. 
Firstly, we describe the exploration process. For each $p_i$, we uniformly sample $m$ candidates and update them with a few iterations. Instead of picking the best one, we keep those candidates as a group $\{ p_{i, u} | u \in [0, m) \}$. Given two poses $p_i$ and $p_j$ where $i \neq j$, we define their \textit{triangular error} as:
\begin{equation}
    l_{i, 0, j}(p_i, p_j) = l_{0, i}(p_i) + l_{0, j}(p_j) + l_{i, j}(\rpose_i + p_j),
\end{equation}
where $\rpose_i + p_j$ is the relative transformation from $x_i$ to $x_j$. The \textit{triangular error} sums up three \textit{pairwise errors}, which indicates the instability of poses $p_i$ and $p_j$ to connect the three views $x_0$, $x_i$, and $x_j$. Then, given a candidate $p_{i, u}$ of $p_i$, we iterate the candidates $p_{j, v}$ of $p_j$ and calculate \textit{triangular errors} ($m$ times). We define the minimum error of $p_{i, u}$ pairing with $p_j$ as:
\begin{equation}
    e_{i, u, j} = \min_{v \in [0, m) } l_{i, 0, j}(p_{i, u}, p_{j, v}).
\end{equation}
We obtain $e_{i, u} = \sum_{j \ne i} e_{i, u, j}$ as the instability of $p_{i, u}$. We select $p_{i, u}$ with lowest instability as the initialization of $p_i$. 
Next, we describe the refinement process. Suppose all poses are initialized, we randomly pick two indices $i$ and $j$ where $0 \le i \neq j < n$. We consider three situations. 
If $i = 0$, we update $p_j$ with the noise error $l(p_j \,;\, x_0, x_j)$.
If $j = 0$, we update $p_i$ with the noise error $l(\rpose_i \,;\, x_i, x_0)$.
If both $i > 0$ and $j > 0$, we will have two poses $p_i$ and $p_j$ to update. We fix one and update the other with the noise error $l(\rpose_i + p_j \,;\, x_i, x_j)$. The refinement process stops upon reaching a preset maximum iteration count.

%-------------------------------------------------------------------------

\section{Experiments}

\subsection{Experimental Settings}
\textbf{Data Preparation.} We conduct experiments on 4 datasets: NAVI \cite{navi}, OmniObject3D \cite{omniobject3d}, Amazon Berkeley Objects (ABO) \cite{abo}, and Common Objects in 3D V2 (CO3D) \cite{co3d}. As the optimization in ID-Pose requires a certain amount of time, we pick a number of samples from each dataset to create testing sets. 

NAVI \cite{navi} is a dataset of casually captured images of 36 category-agnostic objects using hand-held cameras. For each object, we randomly select one set of multi-view images. Each set of images is used as a \textit{view pool}, where we randomly pick out images for pose estimation evaluations.

OmniObject3D \cite{omniobject3d} is a dataset of high-quality real-scanned 3D objects with 190 categories. We select 12 categories ``house'', ``chair'', ``table'', ``bed'', ``ornaments'', ``shoe'', ``scissor'', ``guitar'', ``laundry detergent'', ``dumbbell'', ``dinosaur'', and ``glasses''. For each category, we select the first 5 scanned objects if available. We totally select 54 objects. For each object, we use their Blender rendered images (modified to white background) as a \textit{view pool}.

ABO \cite{abo} is a dataset of 3D synthetic objects of commercial products on Amazon.com. We select 10 objects: ``bag'', ``bar stool chair'', ``ergonomic chair'', ``sofa'', ``couch'', ``desk'', ``shelf'', ``bed'', ``fan'', and ``lamp''. For each object, we render 30 images to create a \textit{view pool}. We place the object at the origin point. We then setup a camera with FOV=$60^{\circ}$ looking at the origin. We randomly move cameras on a sphere to render images. The background is set to white during the rendering.

CO3D (V2) \cite{co3d} is a dataset of turntable-style video sequences captured around real-world objects. It includes 51 object categories. We select 12 categories ``bench'', ``bicycle'', ``car'', ``chair'', ``hydrant'', ``parkingmeter'', ``teddybear'', ``toaster'', ``toybus'', ``toyplane'', ``toytruck'', ``toytrain''. For each category, we select 5 sequences from the test split. For each sequence, we extract every 10 consecutive frames, which are used as a \textit{view pool}.

Given a \textit{view pool}, we pick out images to make testing samples. We define the pick operation as randomly choosing 8 images from the pool. We order the images, and select the first 2, 4, 6, 8 images to make four testing samples, respectively. This eliminates data variants in comparison between using less and more views. For NAVI, we make 210$\times$4 testing samples including 3360 pairs of images to estimate poses. For OmniObject3D, we make 162$\times$4 samples including 2592 pairs. For ABO, we make 100$\times$4 samples including 1600 pairs. For CO3D, we make 180$\times$4 samples including 2880 pairs. Given a sample with $n$ images, we use the first one as the anchor view and $n-1$ target views.

\textbf{Baseline Settings.} We compare ID-Pose with state-of-the-art methods: 1) COLMAP: The Structure-from-Motion pipeline that uses SuperPoint \cite{superpoint} features with LightGlue \cite{lightglue} matching. We use the implementation provided by HLoc \cite{hloc, superglue}. 2) RelPose++ \cite{relpose++}: RelPose++ is the state-of-the-art method that uses network to predict 6DOF camera poses. There are two released model checkpoints. One is trained using full images, the other is trained using masked images where background is removed. We evaluate both of them in the experiments, and separately denote them as FULL and MASK.

\textbf{ID-Pose Settings.} We use the checkpoint of Zero-1-to-3 \cite{zero123} trained with 105000 iterations. We preprocess input images to feature a white background and sufficient margins, thereby aligning with the input distribution of the noise predictor. Given an image, we use a segmentation mask to clear the background and use a window to crop out the object at its center. The size of the window is initially set to the largest bounding box of segmentation masks of all input images. The size is then scaled 1.5x to ensure that there is enough margin in the cropped images. The parameters used in ID-Pose are shown in Table \ref{tab:setting}. They are used across all datasets. We additionally evaluate two ablated variants: 1) ID-Pose Naive: finding poses independently for each pair of the images; 2) ID-Pose TR: exploring poses independently and using triangular relations only during the refinement process.

\begin{table}[t]
  \centering
  \begin{tabular}{@{}l|cc|c@{}}
    \toprule
    & \multicolumn{2}{c|}{Exploration} & Refinement \\
    \midrule
    & Probe (\ref{eq:probe}) & Update & Update \\
    \midrule
    Noise $\epsilon$ & fixed & random & random \\
    Noise step & 0.2 & [0.2, 0.8] & [0.2, 0.8] \\
    Factor $\alpha$ & - & 10.0 & 1.0 \\
    Batch size & 16 & 1 & 1 \\
    Iterations & - & 10 (each pose) & $600 \times (n-1)$ \\
    \bottomrule
  \end{tabular}
  \caption{The parameters used in ID-Pose.}
  \label{tab:setting}
\end{table}

\textbf{Evaluation Settings \& Metrics.} Given $n$ images and the camera pose of a selected anchor view. We estimate the relative poses and obtain the absolute poses for the other $n-1$ images. To obtain the absolute camera positions, we scale the relative position with the radius of the anchor view position. The poses are compared with the ground truth in the world coordinate system. Following RelPose++\cite{relpose++}, we obtain relative rotations of the estimated poses and compute the angular differences to the ground truth. We report the rotation accuracy which is the proportion of the angular differences less than 15 or 30 degree. We also compute the $l_2$ distances from the estimated camera positions to the ground truth positions. The distances are normalized by dividing the corresponding ground truth radius. We report the camera position accuracy which is the proportion of distances less than 20\%.

\subsection{Evaluation Results}
\begin{table*}[t]
  \centering
  \begin{tabular}{@{}l|ccc|ccc|ccc|ccc@{}}
    \toprule
    & \multicolumn{3}{c|}{$n=2$} & \multicolumn{3}{c|}{$n=4$} & \multicolumn{3}{c|}{$n=6$} & \multicolumn{3}{c}{$n=8$} \\
    \midrule
    & $15^{\circ}$ & $30^{\circ}$ & PA & $15^{\circ}$ & $30^{\circ}$ & PA & $15^{\circ}$ & $30^{\circ}$ & PA & $15^{\circ}$ & $30^{\circ}$ & PA\\
    
    \midrule
    
    NAVI & \multicolumn{3}{c|}{Pairs 210} & \multicolumn{3}{c|}{Pairs 630} & \multicolumn{3}{c|}{Pairs 1050} & \multicolumn{3}{c}{Pairs 1470} \\
    \midrule
    COLMAP & 25.24 & 26.67 & 0.00 & 16.35 & 16.51 & 0.16 & 12.67 & 12.76 & 0.10 & 11.09 & 11.16 & 0.41 \\
    $\vdash$ Converge \# & \multicolumn{3}{c|}{ 81 + 0 of 210 } & \multicolumn{3}{c|}{ 32 + 62 of 210 } & \multicolumn{3}{c|}{ 11 + 85 of 210 } & \multicolumn{3}{c}{ 11 + 87 of 210 } \\
    RelPose++ FULL & 38.10 & 63.33 & 21.43 & 36.03 & 62.38 & 21.90 & 35.71 & 61.62 & 22.95 & 36.67 & 62.93 & 22.52 \\
    RelPose++ MASK & 25.24 & 51.43 & 17.14 & 25.08 & 49.52 & 15.87 & 24.10 & 49.71 & 15.90 & 22.45 & 48.37 & 13.88 \\
    ID-Pose Naive & 48.57 & 81.43 & 42.38 & 53.33 & 79.52 & 44.29 & 52.10 & 79.24 & 44.10 & 53.20 & 79.18 & 43.88 \\
    ID-Pose TR & 48.57 & 81.43 & 42.38 & 56.35 & 83.49 & 47.78 & 58.29 & 83.24 & 48.19 & 60.61 & 83.20 & 50.27 \\
    ID-Pose & \textbf{48.57} & \textbf{81.43} & \textbf{42.38} & \textbf{57.30} & \textbf{85.56} & \textbf{48.25} & \textbf{60.19} & \textbf{86.76} & \textbf{50.48} & \textbf{62.52} & \textbf{87.62} & \textbf{51.84} \\
    
    \midrule
    
    OmniObject3D & \multicolumn{3}{c|}{Pairs 162} & \multicolumn{3}{c|}{Pairs 486} & \multicolumn{3}{c|}{Pairs 810} & \multicolumn{3}{c}{Pairs 1134} \\
    \midrule
    COLMAP & 12.35 & 15.43 & 0.00 & 8.23 & 9.88 & 0.00 & 8.27 & 9.63 & 0.37 & 8.73 & 9.88 & 0.88 \\
    $\vdash$ Converge \# & \multicolumn{3}{c|}{ 53 + 0 of 162 } & \multicolumn{3}{c|}{ 20 + 43 of 162 } & \multicolumn{3}{c|}{ 16 + 51 of 162 } & \multicolumn{3}{c}{ 19 + 51 of 162 } \\
    RelPose++ FULL & 12.35 & 30.25 & 8.64 & 9.47 & 27.16 & 7.20 & 9.88 & 27.16 & 8.27 & 10.93 & 30.51 & 9.08 \\
    RelPose++ MASK & 16.05 & 38.27 & 12.96 & 13.37 & 34.36 & 12.76 & 12.72 & 33.83 & 11.60 & 12.17 & 33.33 & 11.46 \\
    ID-Pose Naive & 70.99 & 86.42 & 62.35 & 64.81 & 80.66 & 54.73 & 64.44 & 80.25 & 55.06 & 63.49 & 79.81 & 55.11 \\
    ID-Pose TR & 70.99 & 86.42 & 62.35 & \textbf{69.34} & \textbf{82.10} & \textbf{61.52} & 70.49 & 82.84 & 62.59 & 69.05 & 83.16 & 63.76 \\
    ID-Pose & \textbf{70.99} & \textbf{86.42} & \textbf{62.35} & 67.70 & 81.07 & 59.47 & \textbf{71.23} & \textbf{84.07} & \textbf{64.20} & \textbf{70.11} & \textbf{84.13} & \textbf{64.02} \\
    
    \midrule

    ABO & \multicolumn{3}{c|}{Pairs 100} & \multicolumn{3}{c|}{Pairs 300} & \multicolumn{3}{c|}{Pairs 500} & \multicolumn{3}{c}{Pairs 700} \\
    \midrule
    COLMAP & 5.00 & 7.00 & 0.00 & 2.00 & 2.67 & 0.00 & 1.20 & 1.60 & 0.00 & 1.71 & 2.00 & 0.00 \\
    $\vdash$ Converge \# & \multicolumn{3}{c|}{ 12 + 0 of 100 } & \multicolumn{3}{c|}{ 3 + 10 of 100 } & \multicolumn{3}{c|}{ 0 + 13 of 100 } & \multicolumn{3}{c}{ 2 + 12 of 100 } \\
    RelPose++ FULL & 4.00 & 9.00 & 2.00 & 3.33 & 8.00 & 1.67 & 3.80 & 9.80 & 3.20 & 3.57 & 10.57 & 3.43 \\
    RelPose++ MASK & 3.00 & 8.00 & 1.00 & 2.33 & 9.67 & 1.67 & 4.20 & 9.60 & 4.00 & 4.29 & 10.00 & 3.86 \\
    ID-Pose Naive & 52.00 & 64.00 & 43.00 & 47.67 & 64.33 & 41.33 & 46.60 & 60.80 & 40.80 & 50.71 & 64.86 & 43.86 \\
    ID-Pose TR & 52.00 & 64.00 & 43.00 & \textbf{53.00} & \textbf{64.67} & \textbf{45.33} & 52.80 & 62.00 & 46.80 & 57.86 & 65.71 & 50.29 \\
    ID-Pose & \textbf{52.00} & \textbf{64.00} & \textbf{43.00} & 52.00 & 61.00 & 44.33 & \textbf{55.00} & \textbf{63.40} & \textbf{49.00} & \textbf{59.29} & \textbf{68.00} & \textbf{53.00} \\

    \midrule

    CO3D & \multicolumn{3}{c|}{Pairs 180} & \multicolumn{3}{c|}{Pairs 540} & \multicolumn{3}{c|}{Pairs 900} & \multicolumn{3}{c}{Pairs 1260} \\
    \midrule
    COLMAP & 20.00 & 25.56 & 0.00 & 19.44 & 21.67 & 0.93 & 22.44 & 24.11 & 2.00 & 23.89 & 26.51 & 3.25 \\
    $\vdash$ Converge \#  & \multicolumn{3}{c|}{ 86 + 0 of 180 } & \multicolumn{3}{c|}{ 48 + 54 of 180 } & \multicolumn{3}{c|}{ 54 + 57 of 180 } & \multicolumn{3}{c}{ 60 + 60 of 180 } \\
    RelPose++ FULL & \textbf{83.33} & \textbf{92.22} & \textbf{47.22} & \textbf{84.63} & \textbf{94.44} & 49.44 & \textbf{85.78} & \textbf{94.11} & \textbf{51.11} & \textbf{86.83} & \textbf{94.52} & \textbf{50.24} \\
    RelPose++ MASK & 81.67 & 90.00 & 46.67 & 82.41 & 91.30 & \textbf{49.63} & 81.44 & 91.22 & 49.33 & 82.22 & 91.59 & 47.38 \\
    ID-Pose Naive & 39.44 & 64.44 & 32.22 & 42.22 & 66.30 & 32.59 & 42.00 & 69.67 & 32.78 & 41.51 & 67.30 & 31.83 \\
    ID-Pose TR & 39.44 & 64.44 & 32.22 & 46.30 & 69.07 & 36.48 & 46.44 & 72.78 & 40.56 & 46.59 & 71.11 & 38.65 \\
    ID-Pose & 39.44 & 64.44 & 32.22 & 47.22 & 70.56 & 37.96 & 46.56 & 73.56 & 39.89 & 48.49 & 72.70 & 39.29 \\

    \bottomrule
    
  \end{tabular}
  \caption{Results on the testing samples of all datasets. ``$15^{\circ}$'': rotation accuracy with $15^{\circ}$ threshold. ``$30^{\circ}$'': rotation accuracy with $30^{\circ}$ threshold. ``PA'': camera position accuracy. The values of the results are shown in 100\%. }
  \label{tab:main_result}
\end{table*}

\begin{figure*}[t]
  \centering
    \includegraphics[width=\textwidth]{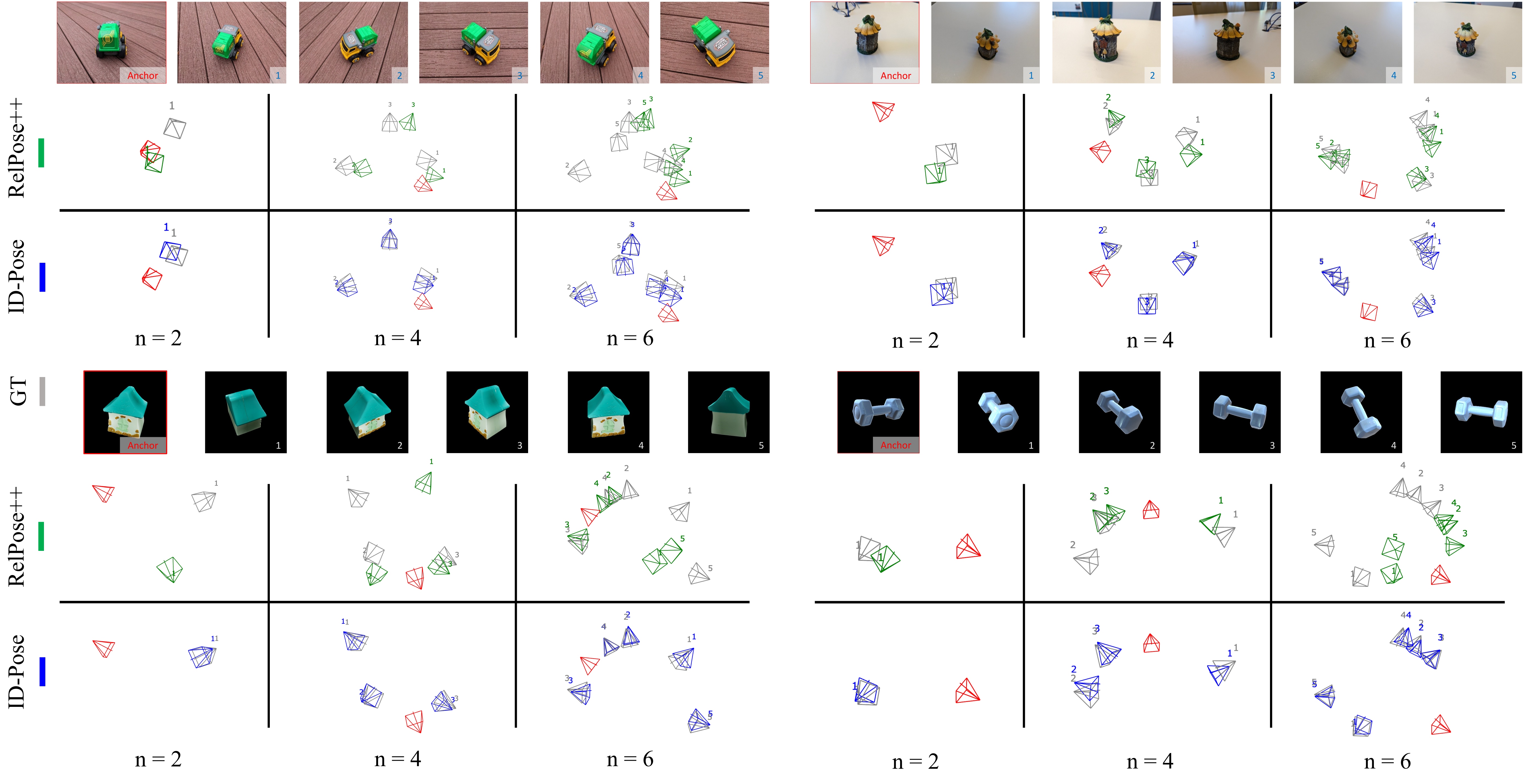}
    \caption{\text{\faSearch} Visualization of the estimated camera poses from the NAVI (top half) and OmniObject3D (bottom half) samples.}
    \label{fig:case1}
\end{figure*}

We report evaluation results aggregated by the number of input views $n$. The results on four datasets are shown in Table \ref{tab:main_result}. 
We compare ID-Pose with the state-of-the-art RelPose++. According to the results on NAVI, OmniObject3D, and ABO samples, ID-Pose significantly outperforms RelPose++ for all different $n$. ID-Pose demonstrates advanced generalization capability on both casually captured photos and rendered images from arbitrary viewpoints. For CO3D samples, ID-Pose shows lower accuracy than RelPose++, which exhibits remarkable performances on this dataset. The reason might be that RelPose++ is trained with CO3D data and the testing samples share a similar distribution. However, as RelPose++ is based on fitting neural networks, the method could fail with out-of-distribution images and camera poses. In comparison, ID-Pose does not require training and generalizes well to a wide range of images.
Note that RelPose++ FULL performs better than MASK on CO3D and NAVI, while performs worse on OmniObject3D and ABO. The reason is that the images of CO3D and NAVI include environmental background. This provides extra information which is learned by FULL to improve accuracy. The images of OmniObject3D and ABO has no background, which are out-of-distribution data for FULL.
We find that COLMAP frequently fails to converge with sparse views, due to lack of reliable features from views with large appearance differences. We additionally report converge counts of running COLMAP. The counts includes 3 numbers A + B of C. A is the count of successful convergences with all $n$ images. B is the count of sub-success cases that the running with all images failed but there exists a success with fewer images. C is the total number of the samples.
The successful count is higher for $n = 2$, as joining two views is simple. With $n > 2$, solving joint matches of multiple sparse views becomes difficult.
We compare ID-Pose with the two ablated variants: Naive and TR. With $n=2$, they show the same results as there is no triangular relation. With $n > 2$ , ID-Pose mostly outperforms the others, and TR outperforms the Naive. This demonstrates the effectiveness of using triangulation. There are some exceptions with $n=4$ for OmniObject3D and ABO, where TR performs the best. As these images are rendered from random viewpoints, there could be abnormal views that affect the exploration process to find promising initial poses.

\begin{figure*}[t]
  \centering
    \includegraphics[width=\textwidth]{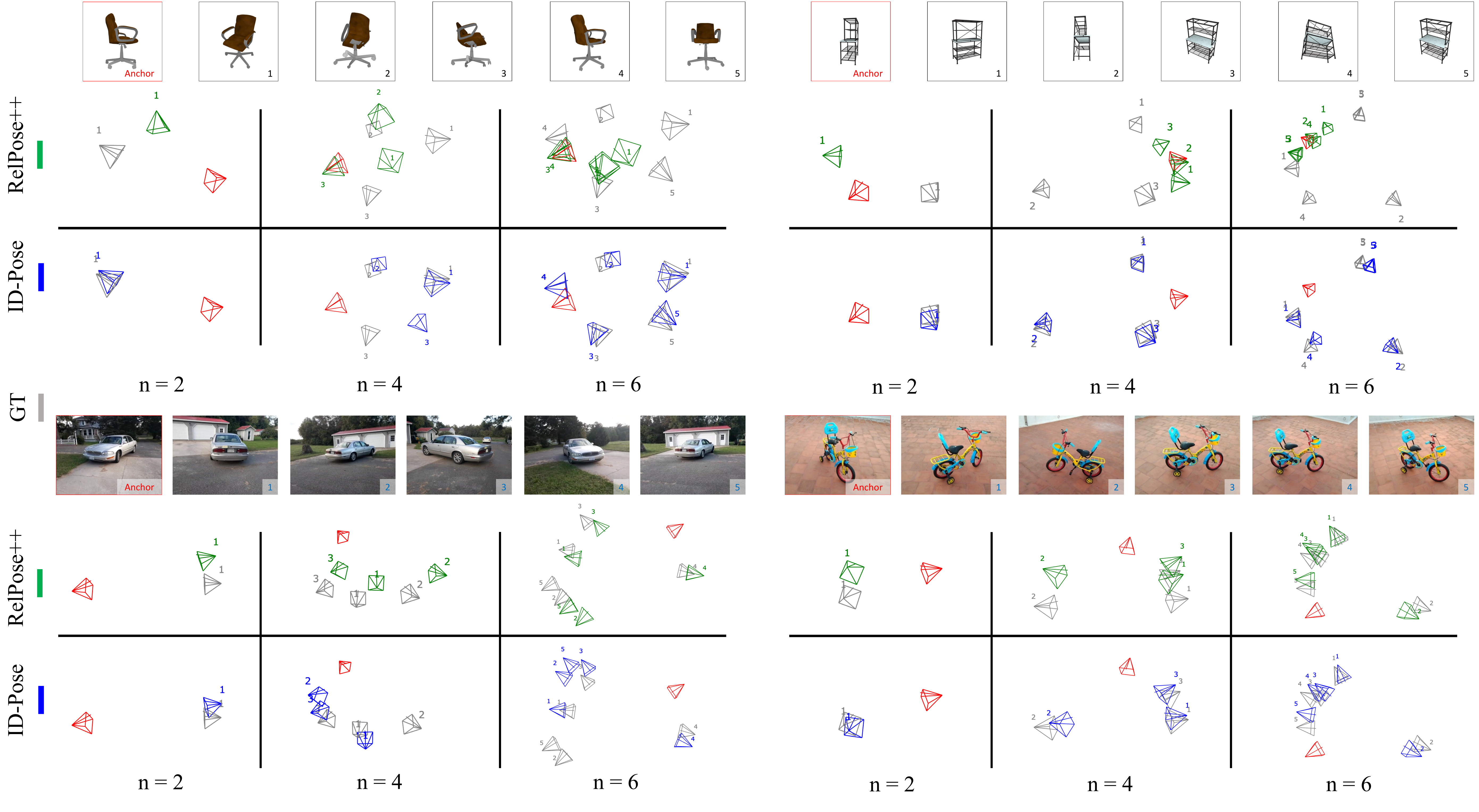}
    \caption{\text{\faSearch} Visualization of the estimated camera poses from the ABO (top half) and CO3D (bottom half) samples.}
    \label{fig:case2}
\end{figure*}

We visualize examples of estimated camera poses in the world coordinate system, under the assumption that camera poses of anchor views are known. We compare ID-Pose with RelPose++ (best version). In Figure \ref{fig:case1}, we show two examples from NAVI and two examples from OmniObject3D, where the cameras are presented by 3D cones. The rotations and shifts between the estimated cones and the ground truth (grey color) can be observed. ID-Pose (blue color) accurately finds the camera poses, and surpasses the performance of RelPose++ (green color). As these objects are category-agnostic, ID-Pose shows better generalization ability due to leveraging the generative model pre-trained on large-scale images. Note that the bottom-right sample is a ``dumbbell'' which exhibits symmetry along its central axis. It is intractable to differentiate two symmetrically opposite views, e.g., image 1 and image 2, based on pure object appearance. However, ID-Pose conditions on lighting to find accurate camera poses. In Figure \ref{fig:case2}, we show two examples from ABO and two examples from CO3D. Given ABO samples, which are synthetic objects, ID-Pose yields accurate results as expected. But RelPose++ fails to predict the poses as it cannot generalize to synthetic content. Given CO3D samples, ID-Pose under-performs RelPose++ which might overfit to this particular data distribution. In general, ID-Pose based on iterative optimization finds more accurate poses than feed-forward prediction. We show some failed cases of ID-Pose. In the ``chair'' example with $n=4$, the pose for image 3 is incorrect. In the ``bicycle'' example with $n=6$, the poses of the image 2 and 5 are incorrect. We find that these prediction are axis-symmetric to the ground truth. This indicates that similar symmetric views can impact the performance of ID-Pose.

\subsection{Noise Step Study}
\begin{figure}[t]
  \centering
    \includegraphics[width=0.9\columnwidth]{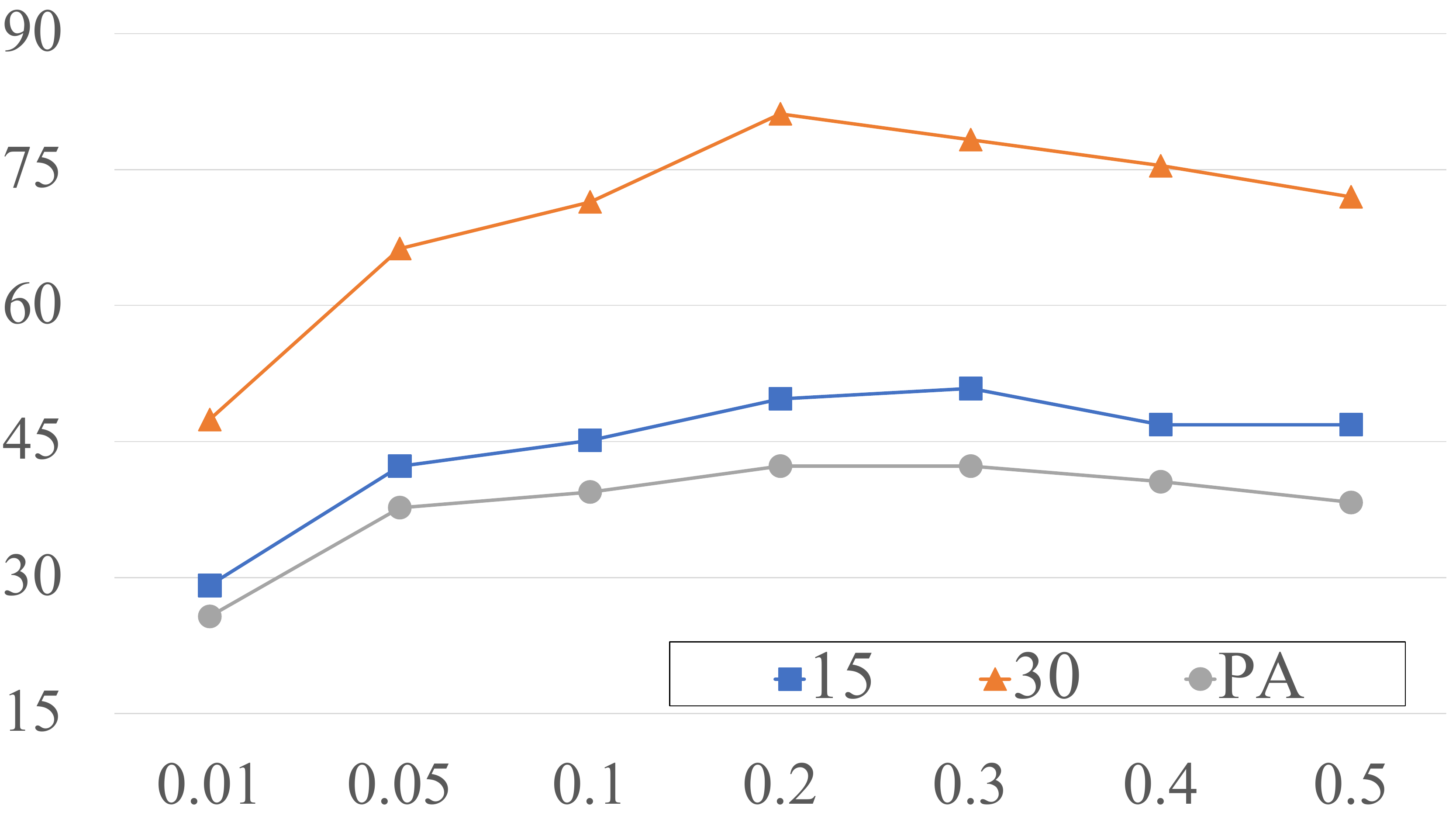}
    \caption{The metrics with respect to noise step $t$.}
    \label{fig:noise}
\end{figure}

We find that choosing the noise step in Equation \eqref{eq:probe} is crucial to the performance of ID-Pose. Because it directly affects the exploration of the initial pose. If the pose is initialized at the wrong side of the sphere, the optimization can hardly drives it to the correct position. To conduct the study, we random pick a number of image pairs from NAVI as a study set. Let $t$ be the noise step, we experiment $t \in \{$0.01, 0.05, 0.1, 0.2, 0.3, 0.4, 0.5$\}$ (ratio of 1000 steps). The results regarding $t$ is plotted in Figure \ref{fig:noise}. ID-Pose achieves optimal performance around $t=0.2 \sim 0.3$, subsequently experiencing a decline. A small $t$ indicates a light noise which leads to a clear view of the target image. If the noise is too light, the gradients could be vanished.

\subsection{Using Self-captured Images}
\begin{figure}[t]
  \centering
    \includegraphics[width=\columnwidth]{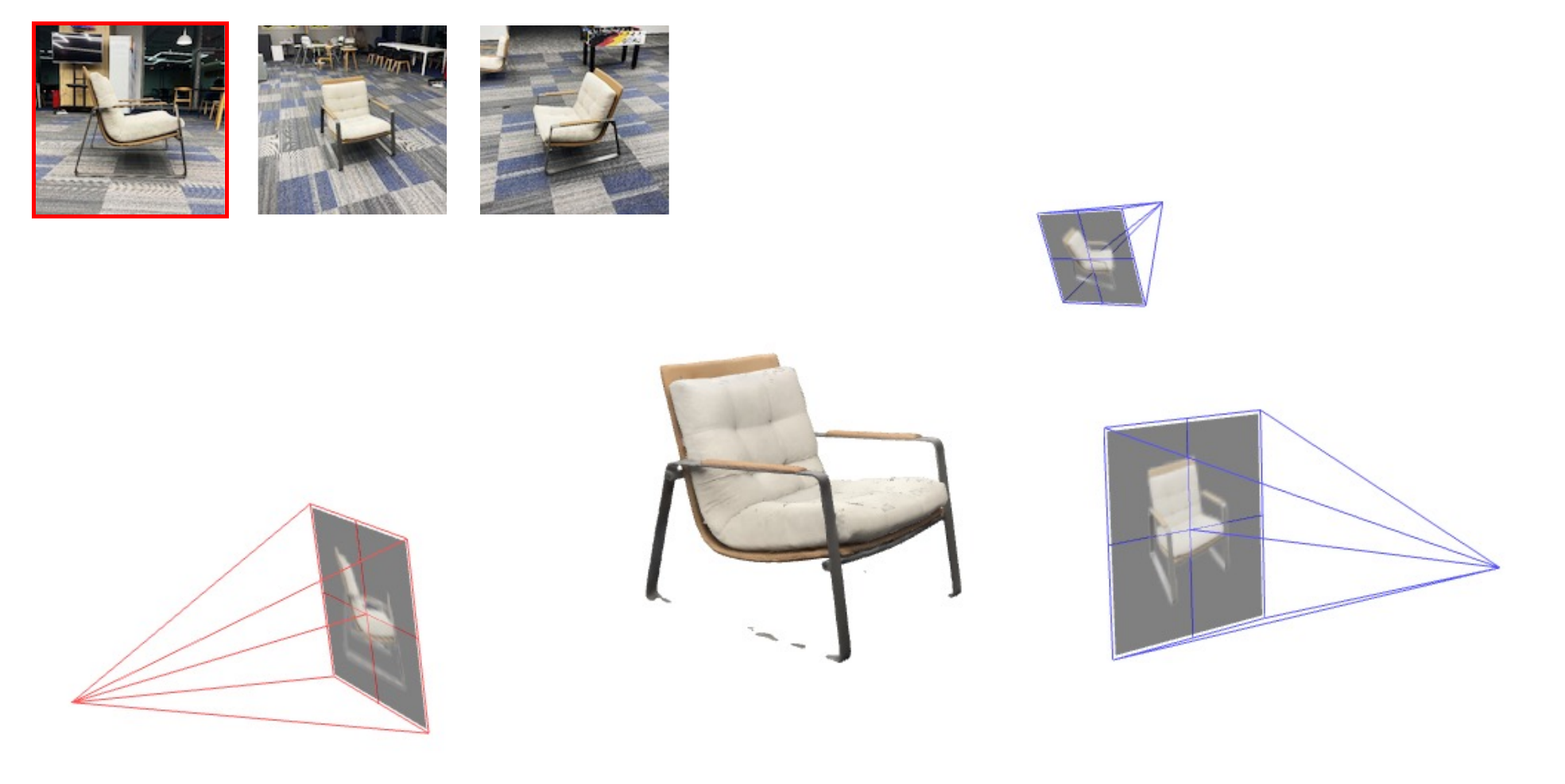}
    \caption{Visualization of the estimated camera poses of the captured images around a chair. The mesh is reconstructed using LumaAI APP. The view in red acts as the anchor view, where the camera pose is manually found using the mesh.}
    \label{fig:sample_daily}
\end{figure}

We qualitatively show the results of ID-Pose on self-captured images. We find four objects in our daily workplace: a toy duck, a toy car, a chair, and a foosball table. We use a smartphone to capture 4-6 images for each of them, and use the DIS method \cite{dis} to segment out the objects with clean background. We select $n$ ($n=2,3,4$) images as input views to estimate the relative poses. To visualize the results, we use the LumaAI APP\footnote{https://lumalabs.ai/} to reconstruct 3D meshes of the objects. We then choose one image as the anchor and manually find its camera pose using the 3D mesh. With the ground truth camera pose of the anchor image, we transform the estimated poses of the target views into the world coordinate system. We show an example of the results of the chair in Figure \ref{fig:sample_daily}. More results are provided in our \href{https://xt4d.github.io/id-pose-web/}{Project Page}.
We observe that the estimated poses cannot align the views to the 3D object perfectly. A reason is that the cameras may not look at the same point in the world, and their ``up'' vectors can be varied. Therefore, pose estimation with the assumption of spherical coordinates will introduce additional errors. 

\subsection{GPU Memory Usage \& Running Time}
We conduct experiments using a single NVIDIA V100 GPU 32G. Running the method requires loading a Zero-1-to-3 checkpoint, where the GPU memory usage is about 28G. The running time includes exploration time and refining time. For two images, the exploration time is about 1 minute. The refining takes about 30 seconds per 100 iterations. Running 600 iterations takes about 3 minutes.

\section{Conclusion \& Limitations}
In this paper, we propose ID-Pose which inverses the denoising diffusion process to estimate camera poses of sparse input views. ID-Pose is a zero-shot method based on the pre-trained Zero-1-to-3 model \cite{zero123}. To find the relative pose of two given images, ID-Pose uses the noise predictor, which conditions on one image and a hypothesis of the relative pose, to predict the noise added on another image. The prediction error is used as the objective function to find the optimal pose with the gradient descent method. ID-Pose can handle more than two images. It estimates each pose with additional pairs of images from triangular relations, where a joint optimization is applied to improve accuracy. We conduct extensive experiments on casually captured photos and rendered images with random viewpoints, where ID-Pose significantly outperforms state-of-the-art methods.

\textbf{Limitations.} The execution time of ID-Pose is considerably greater than that of feed-forward networks. The method requires multiple iterations of pose updating through extensive diffusion networks, which is a computationally costly process. ID-Pose may adopt a more efficient optimization approach to improve the time efficiency. Due to the limitation of Zero-1-to-3, ID-Pose operates under the assumption that all cameras look at the identical point and have no rotation around the optical axis. While the model can accommodate minor deviations, substantial shifts may result in estimation inaccuracies. To mitigate this issue, a potential solution is to fine-tune the novel view diffusion model with the input of 6DOF poses rather than spherical coordinates.

%%%%%%%%% REFERENCES
{\small
\bibliographystyle{ieee_fullname}
\bibliography{paper}

\begin{thebibliography}{10}\itemsep=-1pt

\bibitem{surf}
Herbert Bay, Tinne Tuytelaars, and Luc Van~Gool.
\newblock Surf: Speeded up robust features.
\newblock In {\em ECCV}, page 404–417, 2006.

\bibitem{abo}
Jasmine Collins, Shubham Goel, Kenan Deng, Achleshwar Luthra, Leon Xu, Erhan Gundogdu, Xi Zhang, Tomas~F Yago~Vicente, Thomas Dideriksen, Himanshu Arora, Matthieu Guillaumin, and Jitendra Malik.
\newblock Abo: Dataset and benchmarks for real-world 3d object understanding.
\newblock {\em CVPR}, 2022.

\bibitem{objaverse}
Matt Deitke, Dustin Schwenk, Jordi Salvador, Luca Weihs, Oscar Michel, Eli VanderBilt, Ludwig Schmidt, Kiana Ehsani, Aniruddha Kembhavi, and Ali Farhadi.
\newblock Objaverse: A universe of annotated 3d objects.
\newblock In {\em CVPR}, pages 13142--13153, 2023.

\bibitem{superpoint}
Daniel DeTone, Tomasz Malisiewicz, and Andrew Rabinovich.
\newblock Superpoint: Self-supervised interest point detection and description.
\newblock In {\em CVPRW}, 2018.

\bibitem{textualinversion}
Rinon Gal, Yuval Alaluf, Yuval Atzmon, Or Patashnik, Amit~H Bermano, Gal Chechik, and Daniel Cohen-Or.
\newblock An image is worth one word: Personalizing text-to-image generation using textual inversion.
\newblock {\em arXiv preprint arXiv:2208.01618}, 2022.

\bibitem{diffusion}
Jonathan Ho, Ajay Jain, and Pieter Abbeel.
\newblock Denoising diffusion probabilistic models.
\newblock {\em NeurIPS}, 33:6840--6851, 2020.

\bibitem{navi}
Varun Jampani, Kevis-Kokitsi Maninis, Andreas Engelhardt, Arjun Karpur, Karen Truong, Kyle Sargent, Stefan Popov, Andre Araujo, Ricardo Martin-Brualla, Kaushal Patel, Daniel Vlasic, Vittorio Ferrari, Ameesh Makadia, Ce Liu, Yuanzhen Li, and Howard Zhou.
\newblock Navi: Category-agnostic image collections with high-quality 3d shape and pose annotations.
\newblock 2023.

\bibitem{forge}
Hanwen Jiang, Zhenyu Jiang, Kristen Grauman, and Yuke Zhu.
\newblock Few-view object reconstruction with unknown categories and camera poses.
\newblock {\em ArXiv preprint arXiv:2212.04492}, 2022.

\bibitem{viewformer}
Jon{\'a}{\v{s}} Kulh{\'a}nek, Erik Derner, Torsten Sattler, and Robert Babu{\v{s}}ka.
\newblock Viewformer: Nerf-free neural rendering from few images using transformers.
\newblock In {\em ECCV}, 2022.

\bibitem{relpose++}
Amy Lin, Jason~Y Zhang, Deva Ramanan, and Shubham Tulsiani.
\newblock Relpose++: Recovering 6d poses from sparse-view observations.
\newblock {\em arXiv preprint arXiv:2305.04926}, 2023.

\bibitem{lightglue}
Philipp Lindenberger, Paul-Edouard Sarlin, and Marc Pollefeys.
\newblock Lightglue: Local feature matching at light speed.
\newblock In {\em ICCV}, 2023.

\bibitem{one2345}
Minghua Liu, Chao Xu, Haian Jin, Linghao Chen, Zexiang Xu, Hao Su, et~al.
\newblock One-2-3-45: Any single image to 3d mesh in 45 seconds without per-shape optimization.
\newblock {\em arXiv preprint arXiv:2306.16928}, 2023.

\bibitem{zero123}
Ruoshi Liu, Rundi Wu, Basile Van~Hoorick, Pavel Tokmakov, Sergey Zakharov, and Carl Vondrick.
\newblock Zero-1-to-3: Zero-shot one image to 3d object.
\newblock {\em arXiv preprint arXiv:2303.11328}, 2023.

\bibitem{sift}
David~G. Lowe.
\newblock Distinctive image features from scale-invariant keypoints.
\newblock {\em IJCV}, page 91–110, 2004.

\bibitem{relative}
Iaroslav Melekhov, Juha Ylioinas, Juho Kannala, and Esa Rahtu.
\newblock Relative camera pose estimation using convolutional neural networks.
\newblock In {\em ACIVS}, pages 675--687. Springer, 2017.

\bibitem{nerf}
Ben Mildenhall, Pratul~P. Srinivasan, Matthew Tancik, Jonathan~T. Barron, Ravi Ramamoorthi, and Ren Ng.
\newblock Nerf: Representing scenes as neural radiance fields for view synthesis.
\newblock In {\em ECCV}, 2020.

\bibitem{dbw}
Tom Monnier, Jake Austin, Angjoo Kanazawa, Alexei~A Efros, and Mathieu Aubry.
\newblock Differentiable blocks world: Qualitative 3d decomposition by rendering primitives.
\newblock 2023.

\bibitem{t2i}
Chong Mou, Xintao Wang, Liangbin Xie, Jian Zhang, Zhongang Qi, Ying Shan, and Xiaohu Qie.
\newblock T2i-adapter: Learning adapters to dig out more controllable ability for text-to-image diffusion models.
\newblock {\em arXiv preprint arXiv:2302.08453}, 2023.

\bibitem{dis}
Xuebin Qin, Hang Dai, Xiaobin Hu, Deng-Ping Fan, Ling Shao, and Luc Van~Gool.
\newblock Highly accurate dichotomous image segmentation.
\newblock In {\em ECCV}, 2022.

\bibitem{clip}
Alec Radford, Jong~Wook Kim, Chris Hallacy, Aditya Ramesh, Gabriel Goh, Sandhini Agarwal, Girish Sastry, Amanda Askell, Pamela Mishkin, Jack Clark, et~al.
\newblock Learning transferable visual models from natural language supervision.
\newblock In {\em ICML}, pages 8748--8763. PMLR, 2021.

\bibitem{co3d}
Jeremy Reizenstein, Roman Shapovalov, Philipp Henzler, Luca Sbordone, Patrick Labatut, and David Novotny.
\newblock Common objects in 3d: Large-scale learning and evaluation of real-life 3d category reconstruction.
\newblock In {\em CVPR}, pages 10901--10911, 2021.

\bibitem{latentdiffusion}
Robin Rombach, Andreas Blattmann, Dominik Lorenz, Patrick Esser, and Bj{\"o}rn Ommer.
\newblock High-resolution image synthesis with latent diffusion models.
\newblock In {\em CVPR}, pages 10684--10695, 2022.

\bibitem{srt}
Mehdi~SM Sajjadi, Henning Meyer, Etienne Pot, Urs Bergmann, Klaus Greff, Noha Radwan, Suhani Vora, Mario Lu{\v{c}}i{\'c}, Daniel Duckworth, Alexey Dosovitskiy, et~al.
\newblock Scene representation transformer: Geometry-free novel view synthesis through set-latent scene representations.
\newblock In {\em CVPR}, 2022.

\bibitem{hloc}
Paul-Edouard Sarlin, Cesar Cadena, Roland Siegwart, and Marcin Dymczyk.
\newblock From coarse to fine: Robust hierarchical localization at large scale.
\newblock In {\em CVPR}, 2019.

\bibitem{superglue}
Paul-Edouard Sarlin, Daniel DeTone, Tomasz Malisiewicz, and Andrew Rabinovich.
\newblock Superglue: Learning feature matching with graph neural networks.
\newblock In {\em CVPR}, 2020.

\bibitem{sfm}
Johannes~L. Schonberger and Jan-Michael Frahm.
\newblock Structure-from-motion revisited.
\newblock In {\em CVPR}, 2016.

\bibitem{sparsepose}
Samarth Sinha, Jason~Y Zhang, Andrea Tagliasacchi, Igor Gilitschenski, and David~B Lindell.
\newblock Sparsepose: Sparse-view camera pose regression and refinement.
\newblock In {\em CVPR}, pages 21349--21359, 2023.

\bibitem{daisy}
E. Tola, V. Lepetit, and P. Fua.
\newblock Daisy: An efficient dense descriptor applied to wide-baseline stereo.
\newblock 32(5):815–830, 2010.

\bibitem{mvdepthnet}
Kaixuan Wang and Shaojie Shen.
\newblock Mvdepthnet: Real-time multiview depth estimation neural network.
\newblock In {\em 3DV}, 2018.

\bibitem{omniobject3d}
Tong Wu, Jiarui Zhang, Xiao Fu, Yuxin Wang, Jiawei Ren, Liang Pan, Wayne Wu, Lei Yang, Jiaqi Wang, Chen Qian, et~al.
\newblock Omniobject3d: Large-vocabulary 3d object dataset for realistic perception, reconstruction and generation.
\newblock {\em arXiv preprint arXiv:2301.07525}, 2023.

\bibitem{ganinversion}
Weihao Xia, Yulun Zhang, Yujiu Yang, Jing-Hao Xue, Bolei Zhou, and Ming-Hsuan Yang.
\newblock Gan inversion: A survey.
\newblock {\em IEEE TPAMI}, 45(3):3121--3138, 2022.

\bibitem{c123}
Jianglong Ye, Peng Wang, Kejie Li, Yichun Shi, and Heng Wang.
\newblock Consistent-1-to-3: Consistent image to 3d view synthesis via geometry-aware diffusion models.
\newblock {\em arXiv preprint arXiv:2310.03020}, 2023.

\bibitem{inerf}
Lin Yen-Chen, Pete Florence, Jonathan~T Barron, Alberto Rodriguez, Phillip Isola, and Tsung-Yi Lin.
\newblock inerf: Inverting neural radiance fields for pose estimation.
\newblock In {\em IEEE/RSJ International Conference on Intelligent Robots and Systems (IROS)}, 2021.

\bibitem{pixelnerf}
Alex Yu, Vickie Ye, Matthew Tancik, and Angjoo Kanazawa.
\newblock pixelnerf: Neural radiance fields from one or few images.
\newblock In {\em CVPR}, 2021.

\bibitem{relpose}
Jason~Y Zhang, Deva Ramanan, and Shubham Tulsiani.
\newblock Relpose: Predicting probabilistic relative rotation for single objects in the wild.
\newblock In {\em ECCV}, pages 592--611. Springer, 2022.

\bibitem{controlnet}
Lvmin Zhang and Maneesh Agrawala.
\newblock Adding conditional control to text-to-image diffusion models, 2023.

\bibitem{sparsefusion}
Zhizhuo Zhou and Shubham Tulsiani.
\newblock Sparsefusion: Distilling view-conditioned diffusion for 3d reconstruction.
\newblock In {\em CVPR}, 2023.

\bibitem{junyan}
Jun-Yan Zhu, Philipp Kr{\"a}henb{\"u}hl, Eli Shechtman, and Alexei~A Efros.
\newblock Generative visual manipulation on the natural image manifold.
\newblock In {\em ECCV}, 2016.

\end{thebibliography}
}

\end{document}